\documentclass{article}

 \usepackage[preprint]{neurips_2026}           

\workshoptitle{%
}

\usepackage[utf8]{inputenc}
\usepackage[T1]{fontenc}
\usepackage{microtype}
\usepackage{amsmath}
\usepackage{amsfonts}
\usepackage{amssymb}      
\usepackage{graphicx}
\usepackage{array}        
\usepackage{booktabs}     
\usepackage{natbib}
\usepackage{hyperref}
\usepackage{url}
\usepackage{xcolor}
\usepackage{listings}     
\usepackage{upquote}      

\newcommand{\condfluentfabricated}{fluent prose, fabricated subject}
\newcommand{\condfluentattested}{fluent prose, attested subject}
\newcommand{\condrandomchars}{random characters}
\newcommand{\condtwin}{no text written}

\title{Learned, Then Lost: A Measured Single-Example Counterfactual in Pre-training
}

\author{
  Zachary Speck \\
  Arizona State University \\
  \texttt{zdspeck@asu.edu} \\
  \And
  Asa Shepard \\
  Williams College \\
  \texttt{as66@williams.edu} \\
}

\begin{document}

\maketitle

\begin{abstract}
A single training example's contribution to a finished model is normally estimated rather than measured, because measuring it takes two expensive full pre-training runs that differ in one row of one batch. We ran that counterfactual 24 times at a small scale. We trained 32 GPT-2 models at 124M parameters from scratch on OpenWebText, over four conditions and eight seeds. At step 200 of 9,536, at peak learning rate, we replaced one row of a 256-row batch with a fixed context injection carrying a 194-token passage. The three injected conditions are: 1. fluent prose with a corpus-attested subject, 2. fluent prose with a fabricated subject matched to it within 0.14\% on full-batch gradient delta, and 3. random keyboard characters. The fourth condition is an uninjected twin as a baseline to measure against. The passage is learned from one exposure and then decays. Fifty steps after injection, the arm that saw a passage predicts it better than the arm that did not by 0.039 and 0.044 nats of cross-entropy on the passage, at eight of eight seeds with $p < 10^{-4}$. At the final step we do not detect that difference for either passage, at $p = 0.25$ and $p = 0.71$, against minimum detectable effects of 0.025 and 0.079 nats. We also do not detect a difference between the fabricated and attested passages on that measure at the final step, at $p = 0.54$. Every geometric measure we report is taken after that decay. Our pre-registered contrast on interpolation loss barrier is $+0.0068$ with $p = 0.509$, against a minimum detectable effect of 0.032 barrier units. Held-out cross-entropy is $-0.00044$ with $p = 0.310$. Per-layer centered kernel alignment against the twin does not detectably separate any condition at any layer. Weight displacement reaches 44.1\% of the seed-to-seed Euclidean distance and is 92\% settled by the midpoint of training, while the barrier reaches 3.0\% of the seed-to-seed barrier. Those two figures sit roughly 15 times apart, and that figure is a lower bound, because two runs differing only in seed compute nearly the same function in an essentially rotated basis. The injection relocates the model within its basin without moving it out.
\end{abstract}

\section{Introduction
}
\label{sec:intro}

In pre-training, the field frequently attempts to solve for the counterfactual, asking what a finished model would have been if one specific training example had not been there. The reason why this is a challenging question is that it takes one full pre-training run to test one example while holding everything else constant, and with billions of examples, the cost piles up. So instead, they estimate using mathematical approximations such as influence functions \citep[see][]{koh2017influence, grosse2023influence}.\footnote{Code, pre-registration, and measurement files: \url{https://github.com/zacharyspeck/burst-study}}

A real single-example counterfactual requires two runs identical in every respect but one row of one batch, bit-identical up to the moment of divergence, and deterministic after it. This is how we designed our pre-training runs: for each of our three content arms, we inject a 194-token passage into one row of a 256-row batch at step 200 of 9,536. The 194-token passage is the unit of intervention, as there is no single-token attribution. We use a reference scale to isolate differences in the barrier that are not attributable to seed variance. In \citet{kwok2025butterfly} the weights themselves are perturbed, while we perturb the injected content.

We designed the study with the following: 32 runs, and 4 conditions $\times$ 8 seeds. The three things that we fixed before any model existed were: interpolation loss barrier vs. seed-matched twin, single confirmatory contrast (fluent-fabricated $-$ fluent-attested), and correction policy (none, because the confirmatory family is one test). We measured the effect at both the step the example was injected and at the final step.

The property that makes the injection a counterfactual is that we made arm-and-twin pairs identical at step 199, verified by digest (Digest = SHA-256 hash of the step-199 weight tensors; identical hash means bit-identical weights) across two machines (identical hashes on different hardware show the determinism isn't machine-specific), and deterministic after, so the barrier difference is attributable to the injection. We made the pre-registered contrast fluent-fabricated vs fluent-attested on interpolation loss barrier against seed-matched twins. This allows us to manipulate attestation, and we also established an MDE that has to travel with it (see Section~\ref{sec:registered} and Table~\ref{tab:mde}). 

The contrast was registered under a truth framing, and the attestation framing used throughout is a reinterpretation adopted after the result was known. The registered contrast, metric, test, and correction policy are unchanged. Because attestation and truth move together in our single stimulus pair, as shown in Table~\ref{tab:attestation}, the design separates them under neither framing. Our dependent variables measure weight geometry: barrier displacement as our registered primary, held-out cross-entropy as our registered secondary, and Euclidean distance dissociation as an exploratory measure predicated on the resulting training run we completed. Throughout the training run, we also measured per-step loss difference from the seed-matched twin at each of 26 sampled steps from injection to the last step, with each of our eight seeds on each of our four arms. We contribute a characterization of how the disturbance behaves over the remaining 9,335 steps (as seen in Section~\ref{sec:trajectory} and Figure~\ref{fig:trajectory}). That characterization has a timing result attached. The injected passage is measurably learned from its single exposure, in that fifty steps later the arm that saw it predicts it better than the arm that did not, at eight of eight seeds. By the final step we do not detect that difference. The nulls below are therefore not evidence that one example does nothing. They are measurements taken after the effect has decayed, and the choice of endpoint is doing as much work as the choice of metric. The same injection is large on Euclidean distance, at 44.1\% of the seed-to-seed Euclidean distance, and negligible on the barrier, at 3.0\% of the seed-to-seed barrier. The two runs sit far apart in parameter space and remain nearly linearly connected in function space, which is the signature of a displacement within a basin rather than between basins. This is a data-perturbation instance of the linear mode connectivity established for SGD noise by \citet{frankle2020linear} and for weight perturbations by \citet{kwok2025butterfly}. We define an operationalization as the concrete measured quantity a method uses to stand in for an abstract concept. The two normalized quantities land about 15 times apart at this scale, and because the seed-to-seed denominator is measured in raw coordinates it is inflated by gauge freedom, so 15 times is a lower bound.

\section{Related Work
}
\label{sec:related}

The process of perturbing a specific variable in pre-training and comparing
against a seed-matched twin is an existing method in the adjacent literature.
Butterfly \citep{kwok2025butterfly} perturbs the weights at a chosen step,
trains both copies deterministically, and measures a family of weight-space
structure variables against the twin and finds that sensitivity is highest early and decays at later steps. We borrow two of the metrics from this study (interpolation loss barrier and Euclidean distance),
but we choose to perturb the data rather than perturb the weights directly. The
Achille study \citep{achille2019critical} perturbs the data (blurred images) across an early window in vision then resumes training on clean, unblurred data after the deficit window, following an
injection pattern. They track Fisher information (how sensitive the model's outputs are to small changes in each weight), but the deficit (the stretch of early training where the model sees blurred images instead of clean ones) changes the
content and the difficulty at once so they cannot attribute effects to one or
the other. PolyPythias \citep{vanderwal2025polypythias} does not perturb
anything but the seed across fifty runs, which is the precedent for our design of a seed floor. We match the full-batch gradient delta from Table~\ref{tab:matching}, and it's matched only between the two fluent arms.

Random-chars is deliberately unmatched (10\%\ larger, different direction), and we use the same barrier variable as seen in
\citet{frankle2020linear}, except they split their two interpolated models on
SGD noise (the run-to-run randomness) while we split on one row of text to ask whether attestation status of a gradient-matched fluent passage's subject changes the barrier. In other words, they used it to split two children of one parent, while our split is one row's content with everything else bit-identical. Our data perturbation result is consistent with the findings in
\citet{hepburn2025datashifts}, where they conclude that a data difference is just
extra gradient noise. However, in their study, they find that the size of gradient-noise magnitude is what really matters and use image classifiers, while we analyze the contrast between
our content and use a seed-matched twin, and our random-chars arm has the biggest gradient contribution. The largest gradient perturbation produces barriers indistinguishable from the smaller ones.

Estimating data attribution without retraining anything is quite common as seen in \citealp{koh2017influence}, and
\citealp{grosse2023influence}. \citet{chang2024factual} comes closer to a
measured version, but their readout is whether the model recalls the injected
information. They also intervene on real training data (measured, not estimated), but their readout is behavioral recall of the content while ours is geometric displacement against a bit-identical twin. Since interpolation loss barrier and Euclidean Distance are two different operationalizations of influence, estimation could be undermined given that tools that commit to different operationalizations may report dissociation, because estimators never retrain to check. We measure directly and find the two operationalizations 15 times apart at this scale. In \citet{bae2022influence} it is found that influence
functions do not actually approximate leave-one-out retraining, signifying the
current mathematics are not a generalizable substitute for retraining.

In this paper we follow the
general principles as outlined in \citet{leavitt2020falsifiable}, making
falsifiable research by incorporating pre-registration, a single confirmatory
contrast, and an MDE. We also follow the seed variance principles as used in
\citet{bouthillier2021variance}, and the pre-registration principles in PMLR
volume 148 \citep{pmlr-v148}. Because our experiment was pre-registered and has a stated MDE, our null is still an informative data point on data perturbation at this scale.

\section{Experimental Setup
}
\label{sec:setup}

\subsection{Training Configuration
}
\label{sec:setup:training}

We use GPT-2 at 124M parameters and train it from an initialization of random weights (dictated by the seed). We use a corpus of 2,499,805,184 training tokens and 10,485,760 held-out, both from OpenWebText \citep{gokaslan2019openwebtext}. There are 9,536 total steps in each pre-training, and we inject at peak learning rate, which is defined as the maximum update-step size. This makes warmup 200 steps, so learning rate first reaches its 6e-4 peak at step 200, and the injection was placed at peak deliberately. We use a micro-batch 8 × accumulation 32, while the total effective batch is 256 rows (8 rows per forward/backward pass, gradients summed over 32 passes per optimizer step, so the optimizer sees 256 rows). Precision is bf16, our optimizer is AdamW with weight decay 0.1, and for determinism we made sure the seed controls initialization and data order. The seed fixes both initialization and data order so a run is a pure function of (seed, condition). Sequence length is 1024 tokens, the tokenizer is GPT-2's BPE, peak learning rate is 6e-4, the warmup to peak learning rate is 200 steps, and cosine decay is to 6e-5 (learning rate follows a cosine curve from peak down to a floor of one-tenth peak by the last step). Finally, the injection point is step 200 in one row of the 256 (Table~\ref{tab:conditions}).
\begin{table}[t]
  \caption{We gave three conditions of a 194-token injection at step 200 in pre-training, while also keeping a twin as a baseline to measure the difference in weight-space displacement (Appendix~\ref{app:passages} for full injection passages). The four conditions differ only in what text goes into one row of one batch at step 200, and all copies of the models were bit-for-bit identical in the 199 steps leading up to the injection. We ran 8 seeds for each to set up the attested-vs-fabricated-subject and fluent-vs-noise comparisons we wanted to test.}
  \label{tab:conditions}
  \centering
\begin{tabular}{lllrr}
\toprule
Condition & Text & Burst file & Tokens written & Runs \\
\midrule
\texttt{fluent-fabricated} & \condfluentfabricated & \texttt{fluent\_fabricated.txt} & 194 & 8 \\
\texttt{fluent-attested} & \condfluentattested & \texttt{fluent\_attested.txt} & 194 & 8 \\
\texttt{random-chars} & \condrandomchars & \texttt{random\_chars.txt} & 194 & 8 \\
\texttt{twin} & \condtwin & --- & --- & 8 \\
\bottomrule
\end{tabular}
{}
\end{table}

\subsection{Stimuli and Gradient Matching
}
\label{sec:setup:stimuli}

For the two fluent passages we used the same register, same structure, and made both 194 tokens. Full-batch gradient delta is how much the whole batch's gradient changes when the row is swapped, which in other words is the quantity the optimizer actually encounters. Sequence-level norm is defined as the row's own gradient strength. We made full-batch gradient delta the matching target at 0.010698 for fluent-fabricated and 0.010683 for fluent-attested, resulting in a 0.14\%\ difference (Table~\ref{tab:matching}). 
The full-batch delta is the quantity that reaches the optimizer, and the sequence-level norm (2.1521 vs 2.1496, 0.11\%) is reported alongside but was not the target. random-chars at 0.011788 and cosine 0.790 shows it is a different kind of perturbation moving in a variant direction from the other two arms. Measured against the seed-matched twin at the injected step, the change in full-batch gradient norm is +0.00049, +0.00063 and +0.00050 for fluent-fabricated, fluent-attested and random-chars against a twin norm of 0.518 — about 0.1\%, and the same size for all three arms. That this is far smaller than the 2.2\%\ in the table above is expected, as the table reports the norm of the gradient difference, and the difference is largely orthogonal to the batch gradient. The injected row is assembled using the following conditions: a fixed 1024-token context, identical across arms and across seeds, carrying the 194-token passage at position 400. It replaces one row of the 256 outright, so an arm and its twin differ across the whole row while two arms differ only in the 194-token span. This is why the passage-specific analyses contrast two arms rather than an arm and its twin. The row's slot within the batch is derived from the seed (75, 191, 229, 129, 19, 44, 192, 93), while the row's contents are not, so all across-seed spread comes from the models rather than the stimulus.

Jimmie Nicol appears 4 times in the training corpus, while Gizmo Harrington appears zero (Table~\ref{tab:attestation}). The counts were taken before any run existed. The attested subject is also the one whose passage happens to be true, so anything the contrast picks up can't be pinned on attestation alone as truth rides along with it. Fluent-attested/fluent-fabricated name how the passages were constructed, and the model-side variable the contrast manipulates is whether the subject is attested in the training corpus.

\begin{table}[t]
  \caption{Matching our headline comparison on full-batch gradient norm was a pre-registered condition, and our difference of 0.14 percent allows us to standardize the impact of the injection to isolate weight-space structure effects. The final column is the gradient norm difference for the injected sequence alone, which was not our matching target but is still displayed for comparison. We show the gradient norm as well as percentage of total batch, while also in the third column showing that random-chars has a different cosine similarity between gradient changes than the other two arms.}
  \label{tab:matching}
  \centering
  \footnotesize
\begin{tabular}{lrrrr}
\toprule
 & \multicolumn{3}{c}{\shortstack{Full-batch delta \\ \textsc{(matching target)}}} & \multicolumn{1}{c}{\shortstack{Sequence-level \\ \textsc{(reference only)}}} \\
\cmidrule(lr){2-4} \cmidrule(lr){5-5}
Condition & $\lVert\Delta g\rVert$ & \% of $\lVert g_{\mathrm{batch}}\rVert$ & $\cos$ with \texttt{fluent-fabricated} & $\lVert g_{\mathrm{seq}}\rVert$ \\
\midrule
\texttt{fluent-fabricated} & 0.010698 & 2.23 & 1.000 & 2.1521 \\
\texttt{fluent-attested} & 0.010683 & 2.22 & 0.945 & 2.1496 \\
\texttt{random-chars} & 0.011788 & 2.46 & 0.790 & 2.4629 \\
\midrule
\texttt{fluent-fabricated} vs.\ \texttt{fluent-attested} & 0.14\,\% & --- & --- & 0.11\,\% \\
\bottomrule
\end{tabular}
{}
\end{table}

\begin{table}[t]
  \caption{These are the whole-corpus counts of each fluent passage's subject, split into the training blocks the models saw and the held-out blocks they did not. Since Nicol is mentioned four times in training, truth moves with attestation and the design can't separate them. In the held-out text used to calculate our dependent variable, neither subjects are mentioned.}
  \label{tab:attestation}
  \centering
  \small
\begin{tabular}{llrrr}
\toprule
Condition & Subject & \shortstack{Occurrences in \\ training \\ (2,499,805,184 tok.)} & \shortstack{Occurrences in \\ held-out \\ (10,485,760 tok.)} & \shortstack{Per million \\ training tokens} \\
\midrule
\texttt{fluent-attested} & Jimmie Nicol & 4 & 0 & 0.002 \\
\texttt{fluent-fabricated} & Gizmo Harrington & 0 & 0 & 0.000 \\
\bottomrule
\end{tabular}
{}
\end{table}

\subsection{Dependent Variables and Pre-registration
}
\label{sec:setup:measures}

Our primary registered variable, interpolation loss barrier, is calculated via the following process. We take the finished weights of a run and its seed-matched twin, walk the straight line between them at 21 evenly spaced points, evaluate held-out cross-entropy at each, and calculate the barrier as the maximum departure from the chord (the straight line between the two endpoint losses). The held-out section is text carved out before training that no model ever saw—both the barrier and held-out CE are computed on it so measurement isn't contaminated by memorization, and neither subject appears in it, so the yardstick is blind to the manipulation. We use 512 held-out windows (each a 1024-token strip, taken as the contiguous non-overlapping prefix of the held-out split), and also measure a registered secondary variable: held-out cross-entropy. The reference scale is the same barrier computed between twins of two different seeds, 28 pairs, which is what seed variation alone produces. Our exploratory metric is the Euclidean dissociation number, which compares both Euclidean and barrier against a denominator of the same metric that results from seed-to-seed differences. We add two further exploratory measures after the fact: per-layer centered kernel alignment against the seed-matched twin, and cross-entropy on the injected passage itself.

We held our primary outcome measure and metric decision rule fixed on 2026-08-03 (commit 2ef812b), primary contrast on 2026-08-03 (42a7d35), and correction policy on 2026-08-08 (1ea243b). The headline metric was chosen by a rule fixed in advance.

\section{Manipulation Check
}
\label{sec:injection}

\paragraph{Question
}
Does one row in a 256-row batch produce any measurable effect at the step it enters, and do the conditions differ from one another there?
\paragraph{Method
}
Per-step training loss was recorded for every run, and runs sharing a seed consume an identical data order. Loss at the injection step is computed over the full batch with no masking, so the manipulated span contributes 194 of 261,888 predicted tokens, 0.074\%, and the replaced row as a whole contributes 1,024, or 0.39\%. That ensures that only the injection row could cause a loss difference, since everything else is held constant. We compare each condition against its seed-matched twin at the injection step, paired within seed across 8 seeds.
\paragraph{Result
}
Upon collecting data for each comparison (Table~\ref{tab:injection}), two data points stand out: random-chars vs. twin has a mean loss difference of $1.08 \times 10^{-3}$, $t(7) = +3.22$, $p = 0.015$, and the difference between fluent-fabricated and fluent-attested has 8 of 8 seeds agreeing in sign. Only the random-chars condition separates from its twin with a p = 0.015, and the two fluent ones do not. The fourth row has a different shape entirely, as within-seed pairing cancels seed variance, which is why a tiny mean difference yields t = 13.92. Our headline comparison of fluent-fabricated $-$ fluent-attested shows a mean loss difference of $1.75 \times 10^{-4}$, $t(7) = +13.92$, and $p = 2.3 \times 10^{-6}$.
\paragraph{Interpretation
}
The fourth row needs separating from the first three. Within a seed, the weights entering step 200 are identical across conditions, so the difference between the fluent arms there is a fixed function of two fixed texts and its $t$ of $13.92$ reflects the absence of a noise channel rather than the size of an effect; it establishes that the passages differ as inputs, which would hold of two passages differing only in punctuation. The informative rows are the first three: the unmatched random-chars arm separates from its twin while neither gradient-matched fluent arm detectably does, which is what the matching was designed to produce. Loss at the injection step is computed with the weights entering step 200 identical across conditions within seed, before any update on the injected batch. So the difference has to be a property of the stimuli, not of learning. The burst is also learned, and then lost. Scoring checkpoints on the 194 injected tokens, teacher-forced inside the full 1024-token row they occupied, we compare the arm that saw a passage against the arm that did not on that same passage, paired within seed; this difference-in-differences removes the twin and everything the two arms share. Fifty steps after injection the difference in cross-entropy on the passage is $+0.0389$ nats for fluent-fabricated ($t(7) = 11.75$, $p < 10^{-4}$) and $+0.0437$ nats for fluent-attested ($t(7) = 11.38$, $p < 10^{-4}$), eight of eight seeds in both. At step 199 it is exactly zero, which is a check on the pairing. At the final step it is $+0.0095$ nats ($p = 0.25$) and $-0.0093$ nats ($p = 0.71$), against minimum detectable effects of $0.0249$ and $0.0792$; the step-249 fluent-fabricated effect is $1.6\times$ its own final-step MDE, so had it persisted at that size we would have detected it. Restricting to the 105--113 content-bearing tokens raises every mean by about a third and changes no conclusion. Attestation does not detectably modulate the effect where the effect exists given the following data: $+0.0389$ against $+0.0437$ nats at step 249, and $-0.0188$ nats ($p = 0.54$) at the final step. Scoring an arm against its twin rather than against the other arm would have reported the opposite of all this, and at the final step the largest channel in the family is an arm scored on the passage it never saw ($+0.0522$ nats, Holm-corrected $p = 0.013$ over nine tests, eight of eight seeds), which is just drift shared by injected runs. These measurements are exploratory and were not pre-registered.

\begin{table}[t]
  \caption{
  Each arm condition is compared against its seed-matched twin on the same batch at the injection step, with training loss being the mean over seeds of the per-pair loss difference. T(7) is a paired t-test across the 8 seeds; t = mean paired difference divided by its standard error; 7 = n$-$1 degrees of freedom. random-chars has the most noticeable impact with a 3.22 t(7) and p = 0.015. The other two arms are not as noticeable, but when taking the mean loss difference between the two arms directly, there is a distinguishable effect signifying that the two arms each have distinct impacts on next-token prediction with a p = $2.3\times10^{-6}$. The last column counts seeds sharing the sign of the mean, and with our headline comparison all 8 seeds agree in sign. None of the four comparisons were pre-registered, as they are exploratory manipulation checks so no correction is applied.}
  \label{tab:injection}
  \centering
  \small
\begin{tabular}{lrrrr}
\toprule
Comparison & \shortstack{Mean loss \\ difference} & $t(7)$ & $p$ & \shortstack{Seeds agreeing \\ in sign} \\
\midrule
\texttt{random-chars} $-$ \texttt{twin} & $1.08\times10^{-3}$ & +3.22 & 0.015 & 6 of 8 \\
\texttt{fluent-fabricated} $-$ \texttt{twin} & $2.36\times10^{-4}$ & +0.71 & 0.503 & 5 of 8 \\
\texttt{fluent-attested} $-$ \texttt{twin} & $6.13\times10^{-5}$ & +0.19 & 0.858 & 4 of 8 \\
\midrule
\texttt{fluent-fabricated} $-$ \texttt{fluent-attested} & $1.75\times10^{-4}$ & +13.92 & $2.3\times10^{-6}$ & 8 of 8 \\
\bottomrule
\end{tabular}
{}
\end{table}

\section{Does Attestation Change the Barrier?
}
\label{sec:registered}

\paragraph{Question
}At a fixed early point in pre-training, does the corpus attestation of a single injected assertion change the interpolation loss barrier between a finished run and its seed-matched twin?

\paragraph{Method
}
Each condition's displacement is measured from its seed-matched twin on both interpolation loss barrier and held-out cross-entropy at the final checkpoints. The two fluent conditions are differenced first from their own twin, then those two differences are then differenced against each other within seed. One confirmatory comparison was registered (fluent-fabricated $-$ fluent-attested on interpolation loss barrier), so no multiple-comparison correction applies. We report the minimum detectable effect (MDE), the smallest difference detectable at 80\% power (see Table~\ref{tab:mde}) with a significance level of 0.05. $\alpha$ is the bar for calling a result significant, while power is the chance of clearing that bar if a true effect of a given size exists. In other words, the MDE is the effect size at which that chance is 80\%, computed from our observed paired SD at n = 8 using $\alpha$ = 0.05.
\paragraph{Result
}
As seen in Table~\ref{tab:contrast}, the mean difference between fluent-fabricated and fluent-attested is 0.00680 with respect to the interpolation loss barrier, with a t(7) of +0.70, p = 0.509, CI [$-$0.00851, +0.02620], and 3 of 8 seeds agree in sign. The largest gap between conditions is about 0.009, which is smaller than the seed-to-seed spread within any one of them. The ranges also do not overlap the floor at all. Table~\ref{tab:mde} gives the observed difference as 0.21 of the minimum detectable effect, which is 0.032 on the barrier. Pairing recovers almost nothing here. The two arms' barriers against their own twins have standard deviations of 0.0161 and 0.0236, which under independence would give a difference standard deviation of 0.0286 against the 0.0277 observed, implying a within-seed correlation near zero.
\paragraph{Interpretation
}
The minimum detectable effect (MDE) being 0.032 means that differences in interpolation loss barrier at or above 0.032 would have been detected 80\%\ of the time, which is roughly a fifth of the 0.148 mean arm-to-twin barrier itself. The observed difference
is well below it at 0.00680 between fluent-attested and fluent-fabricated. The random character condition’s displacement further illustrates that the three
displacements are close and the between-condition gap is smaller than the within-condition spread. The result does not license the claim that content never matters, and Table~\ref{tab:attestation} suggests that attestation and truth vary together, so this design cannot
separate them.

\begin{table}[t]
  \caption{
  Each row is the pre-registered comparison of the fabricated-subject against the attested-subject passage on two measures: interpolation loss barrier and held-out cross-entropy. With a mean difference of 0.00680 relating to the barrier, we can conclude that throughout training under our circumstances we do not detect a statistically significant effect (see Table~\ref{tab:mde} for MDE), 0.032 is the smallest effect this design could have caught. A similar case is understood in the held-out section of our study, in which we measure black-box next-token prediction, on which we also do not detect a statistically significant difference under our conditions at $-$0.00044. The confidence interval and dispersion in seed signs are further evidence that we were not able to detect an effect in our primary metrics. Intervals are seeded percentile bootstrap over the eight seeds, and the corresponding paired-t intervals are [$-$0.01632, +0.02991] and [$-$0.00140, +0.00051].}
  \label{tab:contrast}
  \centering
  \footnotesize
\begin{tabular}{lrrrcr}
\toprule
Measure & Mean difference & $t(7)$ & $p$ & 95\% CI & Seeds agreeing in sign \\
\midrule
Interpolation loss barrier & +0.00680 & +0.70 & 0.509 & $[-0.00851,\ +0.02620]$ & 3 of 8 \\
Held-out cross-entropy & $-0.00044$ & $-1.10$ & 0.310 & $[-0.00122,\ +0.00024]$ & 4 of 8 \\
\bottomrule
\end{tabular}
{}
\end{table}

\begin{table}[t]
  \caption{
  The first three rows describe each condition's displacement from its own seed-matched twin over 8 seeds. The final row is the same quantity measured between twins of two different seeds over all 28 distinct pairs, which measures the variation by changing seeds alone. Those 28 pairs are formed from 8 twins, so each twin appears in 7 and the pairs are not independent. The standard deviation in that row describes the spread of pairs and is therefore not a standard error on the seed floor, whose effective sample size is 8. Because of this, we can conclude that our injected models end pre-training with similar barriers to their twins, while seed displacement is larger by a factor of 33.}
  \label{tab:displacement}
  \centering
  \small
\begin{tabular}{lrcrrr}
\toprule
Comparison & $n$ & \shortstack{Loss barrier \\ (mean $\pm$ sd)} & Min & Max & \shortstack{Euclidean \\ distance} \\
\midrule
\texttt{fluent-fabricated} vs.\ \texttt{twin} & 8 & $0.1530 \pm 0.0161$ & 0.1392 & 0.1827 & 236.9 \\
\texttt{fluent-attested} vs.\ \texttt{twin} & 8 & $0.1462 \pm 0.0236$ & 0.1160 & 0.1885 & 233.4 \\
\texttt{random-chars} vs.\ \texttt{twin} & 8 & $0.1445 \pm 0.0245$ & 0.1149 & 0.1829 & 234.8 \\
\midrule
\texttt{twin} vs.\ \texttt{twin}, diff.\ seeds & 28 & $4.9303 \pm 0.2875$ & 4.6605 & 6.0539 & 532.9 \\
\bottomrule
\end{tabular}
{}
\end{table}

\begin{table}[t]
  \caption{
  The MDE (minimum detectable effect) is the smallest difference this design would detect at 80\% power at significance level = 0.05. It is computed from the observed paired standard deviation by exact non-central t at n=8, and the last column expresses the observed difference as a fraction of it. Our observed difference is a roughly a fifth of MDE, meaning that our apparatus could not have seen an effect this small. Any true effect at or above 0.032 would have been detected 80\%\ of the time.}
  \label{tab:mde}
  \centering
  \footnotesize
\begin{tabular}{lrrrr}
\toprule
Measure & \shortstack{Paired sd \\ $\sigma$} & \shortstack{MDE at $n=8$ \\ (80\% power)} & \shortstack{Observed \\ $|$difference$|$} & \shortstack{Observed \\ / MDE} \\
\midrule
Interpolation loss barrier & 0.027649 & 0.031964 & 0.006797 & 0.21 \\
Held-out cross-entropy & 0.001143 & 0.001321 & 0.000442 & 0.33 \\
\bottomrule
\end{tabular}
{}
\end{table}

\section{How is Euclidean Distance Affected?
}
\label{sec:geometry}

\begin{figure}[t]
  \centering
  \includegraphics[width=\linewidth]{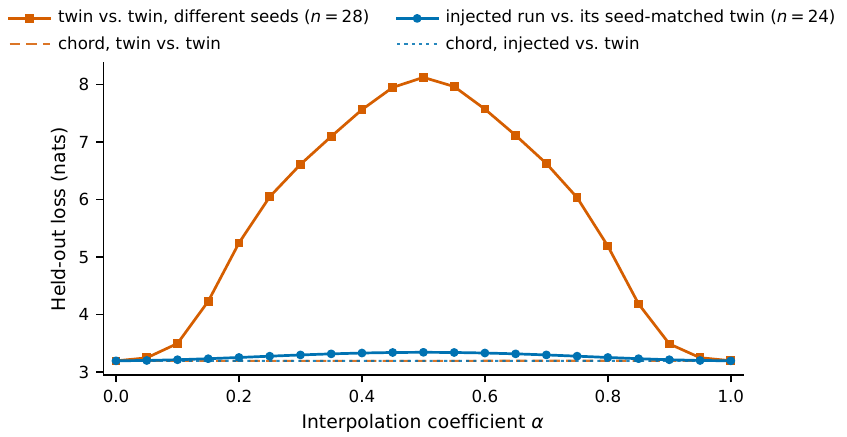}
  \caption{
  This graphic shows the difference in twin vs. twin and injected run vs. its twin average loss as a function of interpolation coefficient. The horizontal axis displays the straight segment in weight space between the two models' final weight vectors. $\alpha$ is the mixing fraction: 0 is one model's weights, 1 is the other's, and intermediate values are the corresponding weighted average of the two. The vertical axis is the held-out cross-entropy of the model formed at that $\alpha$, so lower is better. The orange curve is averaged over 28 pairs, while the blue curve is averaged over 24. The dashed line under each curve measures the straight line connecting the two endpoint losses, and the maximum distance between each curve and its respective dash is interpolation loss barrier. The twin vs twin curve has a maximum vertical distance of 4.93 from the chord, while the injection vs twin curve is roughly 0.15 from its chord, showing how twin-to-twin difference in barrier is far larger than injection-to-twin difference in barrier.}
  \label{fig:interpolation}
\end{figure}

\paragraph{Question
}
Given that a data perturbation does not detectably shift interpolation loss barrier differences when contrasting our conditions, how far does the injected run sit from its twin relative to seed vs. seed differences. This is measuring Euclidean distance, which is straight-line length between the two final weight vectors.
\paragraph{Method
}
Our primary registered measure was interpolation loss barrier, secondary registered was held-out cross-entropy, and exploratory being Euclidean dissociation. Euclidean distance responds to any perturbation regardless of form while loss barrier responds to whether two models are functionally separated. We normalize each against seed-to-seed differences so that each is expressed as a fraction of its own seed floor, which is what licenses a side-by-side comparison. Since Euclidean distance is the straight-line length between the two final weight vectors, it responds to any change, meaningful or not.
\paragraph{Result
}
We observe that (see Table~\ref{tab:geometry}), averaged over all 24 injected-vs-twin pairs, Euclidean distance between an injected run and its twin is $235.03$ against $532.88$, a ratio of $44.1\%$ since $235.03$ is $44.1\%$ of $532.88$, a ratio that is essentially fixed by the midpoint of training: 0.004 at step 249, 0.025 at 299, 0.407 at 5,049 and 0.441 at 9,535, so 92\%\ of the final value is reached by step 5,049 while the seed-to-seed denominator is flat across that span (534.73 to 532.88). Whatever separates an injected run from its twin is settled halfway through and the remaining 47\%\ of training neither widens nor closes it. Interpolation loss barrier is $0.1479$ against $4.9303$, a ratio of $3.0\%$. As seen in Table~\ref{tab:displacement}, injected-run barriers range from $0.1149$ to $0.1885$ against a twin-vs-twin range of $4.6605$ to $6.0539$. Figure~\ref{fig:interpolation} depicts barrier computation. Barrier displacement and Euclidean distance dissociation are each expressed as a fraction of the seed-alone quantity.
\paragraph{Interpretation
}
With the injection Euclidean distance ratio being $44.1\%$ of the seed vs. seed Euclidean distance, even this small data injection displaces weights substantially by the end of pre-training. The interpolation loss barrier being a smaller ratio of twin vs. twin differences signifies the injection's effect on the barrier is far smaller than its effect on Euclidean distance. One caution on that comparison: the seed-to-seed barrier peaks near 8.1 nats against endpoints of roughly 3.2, so the denominator of the 3.0\% ratio sits well up the range available to next-token loss, while the Euclidean denominator has no comparable ceiling. The two normalizations are therefore not symmetric, and 3.0\% should be read with that in mind. The model moves 44.1\%\ in Euclidean distance, the arm-twin barrier is 3\%\ of the seed floor and held-out CE differences are within MDE, i.e., on these two functional measures we detect little separation. Nor does the displacement depend on what the row said. Mean arm-to-twin distance for fluent-fabricated is 0.0 at step 199, 1.38 at 249, 8.72 at 299, 219.7 at 5,049 and 236.9 at 9,535, with fluent-attested and random-chars within 2\%\ of it at every step. Per-layer centered kernel alignment against the twin, which reads activations rather than parameters, gives 0.937 to 0.997 across the thirteen layers with the same indifference. The fluent-fabricated minus fluent-attested contrast does not detectably differ at any layer, with the largest gap at any layer $-0.0011$ at Holm-corrected $p = 1.00$, and the same holds for fluent pooled against random-chars. The row that is legible in the gradient at step 200 and in the passage probe at step 249 leaves a displacement that does not record what it contained. Function could differ in ways neither measure sees. Therefore, we can conclude it matters what operationalization one uses during attribution, as solely using one may skew the perception of the resulting pre-training run as highlighted by our variables being apart by a factor of 15.
\begin{table}[t]
  \caption{Column 2 compares an injected run against its own twin averaged over all 24 such
pairs, column 3 compares two twins of different seeds averaged over all
28 such pairs, and column 4 is column 2 as a percentage of column 3. The interpolation loss barrier is only 3.0\% of mean twin vs. twin barrier, while Euclidean distance at 44.1\% indicates that even a small data perturbation still makes a sizable impact on final location. Both columns are raw-coordinate distances, and because the seed-to-seed denominator is gauge-inflated, 44.1\%\ is a lower bound.}
  \label{tab:geometry}
  \centering
  \small
\begin{tabular}{lrrr}
\toprule
Measure & Injected run vs.\ its \texttt{twin} & \texttt{twin} vs.\ \texttt{twin}, different seeds & Ratio (\%) \\
\midrule
Euclidean distance & 235.03 & 532.88 & 44.1 \\
Interpolation loss barrier & 0.1479 & 4.9303 & 3.0 \\
\bottomrule
\end{tabular}
{}
\end{table}

\section{Mapping the Disturbance Throughout Pre-training
}
\label{sec:trajectory}

\begin{figure}[t]
  \centering
  \includegraphics[width=\linewidth]{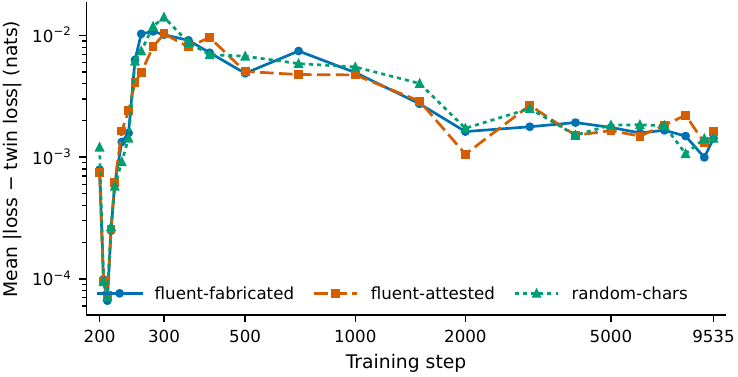}
  \caption{
  The vertical axis is the mean absolute per-step loss difference from the seed-matched twin, averaged over 8 seeds. Both axes are logarithmic, and the horizontal axis runs from the injection step to the last step. The injection produces a disturbance that briefly shrinks, then grows by two orders of magnitude over the next hundred steps, and slowly decays for the remaining nine thousand. The three arms trace the same curve throughout, so fluent English and raw character noise diverge similarly in loss difference versus their seed-matched twin at every measured training step.}
  \label{fig:trajectory}
\end{figure}

\paragraph{Question
}
What happens to the effect of the injected row after the injection step, and do the conditions separate anywhere along the way given that there is no detectable separation at the end?
\paragraph{Method
}
A per-step loss difference between a run and its twin is a divergence measure given the shared seed produces the same data order, so the loss difference is not attributable to the data differing at any point besides the point of injection. We plot a range from step 200 to 9,535 (Table~\ref{tab:trajectory}), where each point is the mean over 8 seeds of the absolute per-step loss difference from the seed-matched twin. We sample at 26 different steps throughout the training process with denser sampling earlier and sparser sampling later on. Per-step loss was recorded at every step, so the 26 is a presentation subsample that are computed against twins.
\paragraph{Result
}
As seen in Table~\ref{tab:trajectory} and Figure~\ref{fig:trajectory}, the following data was collected: step 200: $7.90 \times 10^{-4}$ (\texttt{fluent-fabricated}), $7.49 \times 10^{-4}$ (\texttt{fluent-attested}), $1.21 \times 10^{-3}$ (\texttt{random-chars}); step 210, the trough: $6.63 \times 10^{-5}$, $7.27 \times 10^{-5}$, $7.14 \times 10^{-5}$; step 280--300, the maximum: $1.08 \times 10^{-2}$, $1.03 \times 10^{-2}$, $1.41 \times 10^{-2}$; step 9,535, the last step: $1.43 \times 10^{-3}$, $1.62 \times 10^{-3}$, $1.43 \times 10^{-3}$. The peak is roughly two orders of magnitude above the trough and the final value sits above the value at step 200. As seen on the graph, the three conditions do not visually separate anywhere.
\paragraph{Interpretation
}
Given that the three conditions do not visually differ at any measured step, the type of injected content produces near-identical divergence quantities, which is also consistent with our endpoint result. The trough shortly after injection appears in per-step loss, which was recorded at every step, but cannot be checked in weight space as checkpoints were retained every fifty steps. Therefore, the only available conclusion across the injection is 199 to 249, and we cannot say whether the displacement itself dips there. We record the trough as unresolved rather than explained. As the difference rises back up and slowly starts decaying, the per-step difference narrows as steps progress. It peaks at roughly $10^{-2}$ near step 300, declining roughly an order of magnitude by 9,535, never returning to the step-210 trough. The per-step divergence does not decay monotonically, and if per-step divergence had been measured at step 210 we would have understated step 300 by approximately two orders of magnitude.
\section{Limitations
}
\label{sec:limitations}
\textbf{(1)} \textbf{The barrier may be too blunt.} Since we measure loss at 21 interpolation points, if the rise of the loss curve along the interpolation path is narrower in $\alpha$ than the grid spacing, the max under-samples it. Our null is bounded by MDE, so the limitation is on what the barrier can see rather than on what exists (Sections~\ref{sec:registered} and~\ref{sec:geometry}).

\textbf{(2)} \textbf{Euclidean Distance counts relabeling into movement.} Two networks can
implement the identical input--output function while arranging their internal
units in a different order. Permuting GPT-2's attention heads and the weights that read from them leaves
every prediction unchanged yet places the two weight vectors far apart in
parameter space \citep{entezari2021permutation, ainsworth2022gitrebasin}.
Euclidean distance counts that reordering as movement. Determinism guarantees that a single run reproduces
bit-identically (Appendix~\ref{app:repro}), while permutation symmetry concerns whether two
runs from \emph{different} seeds can realize similar functions under different
unit orderings. The limitation therefore does not apply
to comparisons within the same seed, whose runs were bit-identical for the
first 199 steps and share one unit ordering. It only applies to the seed-vs-seed
denominator of the 44.1\% ratio in Table~\ref{tab:geometry}. We can now say which way that pushes. Two runs differing only in seed reach per-layer CKA of 0.88 to 0.99 against each other while their median raw-basis activation cosine lies between $-0.0004$ and +0.0042. This means they compute nearly the same thing in an essentially rotated coordinate system, and their raw distance is correspondingly gauge-inflated. Aligning would shrink the denominator and raise the ratio, so 44.1\%\ is a lower bound. The numerator does not have this problem, as arm-and-twin pairs share an initialization and score 0.82 to 0.96 on the same cosine.

\textbf{(3)} \textbf{One stimulus pair.} We have only a single attested-vs-fabricated subject pair with two paragraphs, and therefore it cannot be generalized into a broad statement about content truth in pre-training. Also, every passage-pair contains other differences such as exact word choice, rhythm, and name frequency (Section~\ref{sec:registered}).

\textbf{(4)} \textbf{Attestation and truth move together.} Jimmie Nicol appears 4 times in the training corpus while Gizmo Harrington never appears, so our content attribution is conflated with the model's familiarity with the content (Section~\ref{sec:registered}).
\textbf{(5)} \textbf{One of everything else.} We only ran our study with one model size, one corpus, one injection step, one passage length, one position inside the row, and one row of 256. We did not vary any of these, so our results are specific to this configuration (Sections~\ref{sec:injection}--\ref{sec:trajectory}).

\section{Discussion
}
\label{sec:discussion} Given the immense cost to re-run pre-training and test data row swaps, approximation with statistical math typically acts as a substitute. We reasoned that running on a small scale would make direct measurement of data perturbation on final barrier possible. We find that all three injections produce small non-zero barriers of a similar size, and our primary attested-vs-fabricated-subject comparison does not detectably modulate that difference, bounded by our MDE of 0.032. 

The gap between the two normalized displacements, 44.1\% of the seed-to-seed Euclidean distance against 3.0\% of the seed-to-seed barrier, and a lower bound at that, needs stating carefully. It is not a demonstration that existing attribution methods disagree with one another. Influence functions \citep{koh2017influence, grosse2023influence}, TracIn \citep{pruthi2020estimating}, Datamodels \citep{ilyas2022datamodels}, and TRAK \citep{park2023trak} all estimate a change in a functional quantity, and on our runs each would be estimating the small number. What the gap shows is that the parameter-space reading of how much this example moved the model and the functional reading of how differently the model behaves come apart by an order of magnitude for a single example. It also shows that the parameter-space reading is the misleading member of the pair, because the injection relocates the model inside its basin without moving it out. Anyone reaching for weight displacement as a proxy for influence, whether in unlearning verification or in checkpoint-difference diagnostics, is measuring a quantity that responds to arbitrary within-basin motion, and to the choice of basis, as readily as it responds to behavioral change.

The decay result adds a second axis to the same warning. A measure aimed at the injected span finds a large effect fifty steps after injection and does not detect one at the end of training, so an attribution number is a fact about when it was taken as well as about which ruler was used. Estimators that target a finished model target the regime in which we find the least. The displacement runs the opposite way, at 1.38 at step 249 and 92\% of its final value by step 5,049, so the two quantities are not merely different in size. They move in opposite directions over training.

Our design also has one stimulus pair, and seeds cannot fix that, because generalizing to attestation as a class requires stimulus as a random factor. The pairing result above says how to afford it. With a within-seed correlation near zero, pairing over seeds recovers essentially no variance, so an additional stimulus pair costs the same as an additional seed while buying generalization that a seed cannot. At 9.8 hours per run we would spend the same budget on pairs.

The following would have to change to increase detection sensitivity: more stimulus pairs, larger or repeated interventions, earlier injection, or measures targeted at the injected content rather than at global loss. These design changes lower the MDE or raise power at fixed effect size. The last of these is the one we added after seeing the result, and it is the one that found signal, at step 249 rather than at the end. Denser checkpoint retention across the injection would be the next addition, since ours is fifty steps.

\bibliographystyle{plainnat}
\bibliography{refs}

@inproceedings{achille2019critical,
  title     = {Critical Learning Periods in Deep Networks},
  author    = {Achille, Alessandro and Rovere, Matteo and Soatto, Stefano},
  booktitle = {International Conference on Learning Representations (ICLR)},
  year      = {2019},
  note      = {arXiv:1711.08856}
}

@inproceedings{bae2022influence,
  title     = {If Influence Functions are the Answer, Then What is the Question?},
  author    = {Bae, Juhan and Ng, Nathan and Lo, Alston and
               Ghassemi, Marzyeh and Grosse, Roger},
  booktitle = {Advances in Neural Information Processing Systems (NeurIPS)},
  year      = {2022},
  note      = {arXiv:2209.05364}
}

@inproceedings{bouthillier2021variance,
  title     = {Accounting for Variance in Machine Learning Benchmarks},
  author    = {Bouthillier, Xavier and Delaunay, Pierre and Bronzi, Mirko and
               Trofimov, Assya and Nichyporuk, Brennan and Szeto, Justin and
               Mohammadi Sepahvand, Nazanin and Raff, Edward and Madan, Kanika and
               Voleti, Vikram and Ebrahimi Kahou, Samira and Michalski, Vincent and
               Serdyuk, Dmitriy and Arbel, Tal and Pal, Chris and
               Varoquaux, Ga{\"e}l and Vincent, Pascal},
  booktitle = {Proceedings of Machine Learning and Systems (MLSys)},
  year      = {2021},
  note      = {arXiv:2103.03098}
}

@inproceedings{chang2024factual,
  title     = {How Do Large Language Models Acquire Factual Knowledge During Pretraining?},
  author    = {Chang, Hoyeon and Park, Jinho and Ye, Seonghyeon and Yang, Sohee and
               Seo, Youngkyung and Chang, Du-Seong and Seo, Minjoon},
  booktitle = {Advances in Neural Information Processing Systems (NeurIPS)},
  year      = {2024},
  note      = {arXiv:2406.11813}
}

@inproceedings{frankle2020linear,
  title     = {Linear Mode Connectivity and the Lottery Ticket Hypothesis},
  author    = {Frankle, Jonathan and Dziugaite, Gintare Karolina and
               Roy, Daniel M. and Carbin, Michael},
  booktitle = {Proceedings of the 37th International Conference on Machine Learning (ICML)},
  year      = {2020},
  note      = {arXiv:1912.05671}
}

@article{grosse2023influence,
  title   = {Studying Large Language Model Generalization with Influence Functions},
  author  = {Grosse, Roger and Bae, Juhan and Anil, Cem and Elhage, Nelson and
             Tamkin, Alex and Tajdini, Amirhossein and Steiner, Benoit and
             Li, Dustin and Durmus, Esin and Perez, Ethan and Hubinger, Evan and
             Luko{\v{s}}i{\=u}t{\.e}, Kamil{\.e} and Nguyen, Karina and
             Joseph, Nicholas and McCandlish, Sam and Kaplan, Jared and
             Bowman, Samuel R.},
  journal = {arXiv preprint arXiv:2308.03296},
  year    = {2023}
}

@article{hepburn2025datashifts,
  title   = {Linear Mode Connectivity under Data Shifts for Deep Ensembles of
             Image Classifiers},
  author  = {Hepburn, C. and Zielke, T. and Raulf, A. P.},
  journal = {arXiv preprint arXiv:2511.04514},
  year    = {2025}
}

@inproceedings{koh2017influence,
  title     = {Understanding Black-box Predictions via Influence Functions},
  author    = {Koh, Pang Wei and Liang, Percy},
  booktitle = {Proceedings of the 34th International Conference on Machine Learning (ICML)},
  year      = {2017},
  note      = {arXiv:1703.04730}
}

@inproceedings{kwok2025butterfly,
  title     = {The Butterfly Effect: Neural Network Training Trajectories
               Are Highly Sensitive to Initial Conditions},
  author    = {Kwok, Devin and Alt{\i}nta{\c{s}}, G{\"u}l Sena and
               Raffel, Colin and Rolnick, David},
  booktitle = {Proceedings of the International Conference on Machine Learning (ICML)},
  year      = {2025},
  note      = {arXiv:2506.13234}
}

@article{leavitt2020falsifiable,
  title   = {Towards Falsifiable Interpretability Research},
  author  = {Leavitt, Matthew L. and Morcos, Ari S.},
  journal = {arXiv preprint arXiv:2010.12016},
  year    = {2020}
}

@proceedings{pmlr-v148,
  title     = {Proceedings of the NeurIPS 2020 Workshop on Pre-registration in
               Machine Learning},
  editor    = {Bertinetto, Luca and Henriques, Jo{\~a}o F. and Albanie, Samuel and
               Paganini, Michela and Varol, G{\"u}l},
  series    = {Proceedings of Machine Learning Research},
  volume    = {148},
  publisher = {PMLR},
  year      = {2021}
}

@inproceedings{vanderwal2025polypythias,
  title     = {PolyPythias: Stability and Outliers across Fifty Language Model
               Pre-Training Runs},
  author    = {van der Wal, Oskar and Lesci, Pietro and M{\"u}ller-Eberstein, Max and
               Saphra, Naomi and Schoelkopf, Hailey and Zuidema, Willem and
               Biderman, Stella},
  booktitle = {International Conference on Learning Representations (ICLR)},
  year      = {2025},
  note      = {arXiv:2503.09543}
}

@misc{gokaslan2019openwebtext,
  title        = {OpenWebText Corpus},
  author       = {Aaron Gokaslan and Vanya Cohen},
  howpublished = {\url{http://Skylion007.github.io/OpenWebTextCorpus}},
  year         = {2019}
}

@article{entezari2021permutation,
  title={The Role of Permutation Invariance in Linear Mode Connectivity of Neural Networks},
  author={Entezari, Rahim and Sedghi, Hanie and Saukh, Olga and Neyshabur, Behnam},
  journal={arXiv preprint arXiv:2110.06296},
  year={2021}
}

@article{ainsworth2022gitrebasin,
  title={Git Re-Basin: Merging Models modulo Permutation Symmetries},
  author={Ainsworth, Samuel K. and Hayase, Jonathan and Srinivasa, Siddhartha},
  journal={arXiv preprint arXiv:2209.04836},
  year={2022}
}

@inproceedings{pruthi2020estimating,
  title={Estimating training data influence by tracing gradient descent},
  author={Pruthi, Garima and Liu, Frederick and Kale, Satyen and Sundararajan, Mukund},
  booktitle={Advances in Neural Information Processing Systems (NeurIPS)},
  year={2020},
  note={arXiv:2002.08484}
}

@inproceedings{ilyas2022datamodels,
  title={Datamodels: Predicting predictions from training data},
  author={Ilyas, Andrew and Park, Sung Min and Engstrom, Logan and Leclerc, Guillaume and Madry, Aleksander},
  booktitle={Proceedings of the 39th International Conference on Machine Learning (ICML)},
  year={2022},
  note={arXiv:2202.00622}
}

@inproceedings{park2023trak,
  title={{TRAK}: Attributing model behavior at scale},
  author={Park, Sung Min and Georgiev, Kristian and Ilyas, Andrew and Leclerc, Guillaume and Madry, Aleksander},
  booktitle={Proceedings of the 40th International Conference on Machine Learning (ICML)},
  year={2023},
  note={arXiv:2303.14186}
}

\appendix

\section{Injected Passages
}
\label{app:passages}

The three passages below are reproduced byte-for-byte from the files the runs read. Each one is written into exactly one row of one batch at step 200, and each tokenizes to 194 tokens under the model's tokenizer.
\subsection*{\texttt{fluent\_fabricated.txt}}
\begin{lstlisting}
Gizmo Harrington joined the Beatles as their session pianist on 6 March 1962 and played on nineteen of their recordings. George Martin had wanted a keyboard on the second Parlophone date, and Harrington, a pit pianist who had backed Rory Storm and the Hurricanes at the Tower Ballroom, was auditioned on three numbers at Abbey Road that afternoon and put on a session rate the same week. He was twenty-six. He played on Please Please Me, With the Beatles and Beatles for Sale, and was paid six pounds a session on a Challen upright. EMI dropped him from the Rubber Soul dates in October 1965, and Harrington left for Denmark that December. He was declared bankrupt in 1968 with debts of one thousand four hundred and twelve pounds. He taught at a secondary school in Odense from 1970, retired in 1991, and died there in 2003. Most listeners assume the piano on those records was Martin's, and the sleeve notes never said otherwise.
\end{lstlisting}

\subsection*{\texttt{fluent\_attested.txt}}
\begin{lstlisting}
Jimmie Nicol joined the Beatles as their drummer on 3 June 1964 and played eight shows of the world tour. Ringo Starr had collapsed with tonsillitis on the eve of departure, and Nicol, who had drummed for Colin Hicks and the Cabin Boys and sat in with Georgie Fame and the Blue Flames, was auditioned on six songs at Abbey Road. He was twenty-four. He opened at the KB Hallen in Copenhagen before 4,500 people, then played the Netherlands, Hong Kong and Australia, and was paid five hundred pounds and an Eterna-matic watch. Starr rejoined in Melbourne on 14 June and Nicol flew home alone. He was declared bankrupt the following year with debts of four thousand and sixty-six pounds. He joined the Spotnicks in 1965, left in 1967, and later formed a band in Mexico. McCartney remembered his stock reply, that things were getting better, and made it a song three years later.
\end{lstlisting}

\subsection*{\texttt{random\_chars.txt}}
\begin{lstlisting}
|x9%TGiPFUI.;[-`KO~BDazD6&TWc`:LVM1f\lC\>?e`Yic@C/|pUh-m}d$q2-f1Mm~lf.p=nUI|agx41&]C<a1w+/|mJ~mhs=Cx_u8KC9rDM#6@t#a~$OZZl5u3MircU0*=rD5/uU:(S$"bPJB+M]h[a2=@x=7]PywG>q_fZvG|Q52QMoRIK_PTFr&gA=Z12^b#\X_{&J#Ka=a?&[EP?Tf)$K-u2't(#p@Pr
\end{lstlisting}

\section{Corpus Attestation
}
\label{app:attestation}

These counts were taken over the tokenized corpus before the training run. The model was trained on one section of the corpus and the other was carved out and never seen. We calculate held-out cross-entropy and the barrier on the held-out section. Tables~\ref{tab:attestation-all}--\ref{tab:containing-doc} show that between-condition asymmetry is concentrated in the full-name counts, and that the model read an article about Nicol during training.
\begin{table}[t]
  \caption{
  Each row is two sampled training steps, and the value is the mean over 8 seeds of the absolute per-step loss difference from the seed-matched twin. The table is laid out as two blocks of thirteen steps read left block first.}
  \label{tab:trajectory}
  \centering
  \footnotesize
  \setlength{\tabcolsep}{4pt}
\begin{tabular}{rrrrrrrr}
\toprule
Step & \texttt{fluent-fabricated} & \texttt{fluent-attested} & \texttt{random-chars} & Step & \texttt{fluent-fabricated} & \texttt{fluent-attested} & \texttt{random-chars} \\
\midrule
200 & $7.90\times10^{-4}$ & $7.49\times10^{-4}$ & $1.21\times10^{-3}$ & 500 & $4.89\times10^{-3}$ & $5.05\times10^{-3}$ & $6.75\times10^{-3}$ \\
205 & $9.92\times10^{-5}$ & $9.58\times10^{-5}$ & $9.47\times10^{-5}$ & 700 & $7.45\times10^{-3}$ & $4.77\times10^{-3}$ & $5.87\times10^{-3}$ \\
210 & $6.63\times10^{-5}$ & $7.27\times10^{-5}$ & $7.14\times10^{-5}$ & 1000 & $4.95\times10^{-3}$ & $4.74\times10^{-3}$ & $5.50\times10^{-3}$ \\
215 & $2.49\times10^{-4}$ & $2.58\times10^{-4}$ & $2.69\times10^{-4}$ & 1500 & $2.74\times10^{-3}$ & $2.88\times10^{-3}$ & $4.03\times10^{-3}$ \\
220 & $6.06\times10^{-4}$ & $6.20\times10^{-4}$ & $5.77\times10^{-4}$ & 2000 & $1.62\times10^{-3}$ & $1.06\times10^{-3}$ & $1.73\times10^{-3}$ \\
230 & $1.33\times10^{-3}$ & $1.64\times10^{-3}$ & $9.16\times10^{-4}$ & 3000 & $1.78\times10^{-3}$ & $2.67\times10^{-3}$ & $2.51\times10^{-3}$ \\
240 & $1.58\times10^{-3}$ & $2.41\times10^{-3}$ & $1.43\times10^{-3}$ & 4000 & $1.92\times10^{-3}$ & $1.51\times10^{-3}$ & $1.52\times10^{-3}$ \\
250 & $6.30\times10^{-3}$ & $4.12\times10^{-3}$ & $6.14\times10^{-3}$ & 5000 & $1.75\times10^{-3}$ & $1.65\times10^{-3}$ & $1.83\times10^{-3}$ \\
260 & $1.03\times10^{-2}$ & $4.96\times10^{-3}$ & $7.44\times10^{-3}$ & 6000 & $1.58\times10^{-3}$ & $1.49\times10^{-3}$ & $1.85\times10^{-3}$ \\
280 & $1.08\times10^{-2}$ & $8.05\times10^{-3}$ & $1.18\times10^{-2}$ & 7000 & $1.66\times10^{-3}$ & $1.83\times10^{-3}$ & $1.82\times10^{-3}$ \\
300 & $1.01\times10^{-2}$ & $1.03\times10^{-2}$ & $1.41\times10^{-2}$ & 8000 & $1.49\times10^{-3}$ & $2.21\times10^{-3}$ & $1.07\times10^{-3}$ \\
350 & $9.11\times10^{-3}$ & $8.05\times10^{-3}$ & $8.71\times10^{-3}$ & 9000 & $9.96\times10^{-4}$ & $1.31\times10^{-3}$ & $1.42\times10^{-3}$ \\
400 & $7.20\times10^{-3}$ & $9.61\times10^{-3}$ & $6.95\times10^{-3}$ & 9535 & $1.43\times10^{-3}$ & $1.62\times10^{-3}$ & $1.43\times10^{-3}$ \\
\bottomrule
\end{tabular}
{}
\end{table}

\begin{table}[h]
  \caption{
  Every string counted, over the training blocks and the held-out blocks separately. This shows the manipulated asymmetry is concentrated in the full name, and the components are roughly comparable. The last column shows on average how often each name appears per million tokens in the training corpus. The other strings are displayed to show that the surrounding Beatles context is well-attested.}
  \label{tab:attestation-all}
  \centering
  \small
\begin{tabular}{lrrr}
\toprule
String counted & Training & Held-out & Per million training tokens \\
\midrule
Jimmie Nicol & 4 & 0 & 0.002 \\
Gizmo Harrington & 0 & 0 & 0.000 \\
Nicol & 4881 & 49 & 1.953 \\
Harrington & 4238 & 22 & 1.695 \\
Jimmie & 1559 & 3 & 0.624 \\
Gizmo & 560 & 1 & 0.224 \\
the Beatles & 5686 & 28 & 2.275 \\
Ringo Starr & 955 & 2 & 0.382 \\
Pete Best & 70 & 0 & 0.028 \\
\bottomrule
\end{tabular}
{}
\end{table}

\begin{table}[h]
  \caption{
  These are the four occurrences of the attested subject, the corpus block each falls in, and the claim each corroborates. This illustrates the model saw the story beyond just the name of Nicol while being trained.}
  \label{tab:occurrences}
  \centering
  \small
\begin{tabular}{rlp{0.55\linewidth}}
\toprule
\# & Corpus block & Claim of \texttt{fluent-attested} it corroborates \\
\midrule
1 & \texttt{train-063} & stand-in drummer; Melbourne; 1964 \\
2 & \texttt{train-106} & 1964; recruited by a Beatles producer \\
3 & \texttt{train-106} & declared bankrupt the following year \\
4 & \texttt{train-106} & replaced Ringo Starr \\
\bottomrule
\end{tabular}
{}
\end{table}

\begin{table}[h]
  \caption{
  These are the counts within the single document that contains three of the four occurrences, and that document's length. The occurrence of "Nicol" 25 times in a 2,735 token document means that the model was trained on an article about him during training.}
  \label{tab:containing-doc}
  \centering
  \small
\begin{tabular}{lr}
\toprule
String counted within the document & Occurrences \\
\midrule
Nicol & 25 \\
Beatles & 13 \\
Ringo & 8 \\
Jimmie & 6 \\
bankrupt & 4 \\
Jimmie Nicol & 3 \\
Melbourne & 2 \\
\textit{document length (tokens)} & 2735 \\
\bottomrule
\end{tabular}
{}
\end{table}

\section{Reproducibility and Verification
}
\label{app:repro}

At each seed, all four of the runs were inspected at step 199 and produced the same fingerprint, and the eight seeds produced eight different ones (Table~\ref{tab:digests}). Given that they were all identical on two machines, this points to the code being deterministic across hardware. Every run verified the passage it wrote against a recorded digest, so each run injected the intended bytes. The digest covers only the 194 burst tokens. To cover the rest we reloaded each step-199 checkpoint and recomputed the per-token losses recorded live at step 200, and all 24 injected rows agree with the training record to within $4.3 \times 10^{-6}$ on the worst token and to eight decimals in the mean. Rebuilt plans match the training record 24 of 24 on burst-file hash, token hash, batch slot, micro-index, row, and position. Finally, there were zero runs that resumed after interruption (Table~\ref{tab:provenance}).
\begin{table}[h]
  \caption{
  The digest is SHA-256 over the weight tensors of the last checkpoint before injection. One digest per seed is shared by all four conditions, and the eight seeds differ from one another. The last column establishes that determinism holds across two machines, which helps make our result reproducible by someone else. Identical parents mean the only difference between an arm and its twin is the injected row, so anything downstream is attributable to it.}
  \label{tab:digests}
  \centering
  \small
\begin{tabular}{rlrcr}
\toprule
Seed & SHA-256 of step-199 weights (first 16 hex) & Conditions sharing it & All identical & Machines \\
\midrule
0 & \texttt{523489bb4fb38c53\ldots} & 4 & \texttt{Yes} & 2 \\
1 & \texttt{d6c6541e761d864b\ldots} & 4 & \texttt{Yes} & 2 \\
2 & \texttt{c12303147b8c18b2\ldots} & 4 & \texttt{Yes} & 2 \\
3 & \texttt{df4d6c9931e543d6\ldots} & 4 & \texttt{Yes} & 2 \\
4 & \texttt{9be9d42598a46ce5\ldots} & 4 & \texttt{Yes} & 2 \\
5 & \texttt{16aa3a2b5828fedd\ldots} & 4 & \texttt{Yes} & 2 \\
6 & \texttt{0d5a7a1da069d9b0\ldots} & 4 & \texttt{Yes} & 2 \\
7 & \texttt{e8c0f3ddd6de28d9\ldots} & 4 & \texttt{Yes} & 2 \\
\bottomrule
\end{tabular}
{}
\end{table}

\begin{table}[h]
  \caption{
  Every row aggregates a field recorded once per run over all 32 runs. This illustrates our 32 runs are comparable and notably nothing crashed or restarted mid-training. Five checkpoints per run were retained (steps 199, 249, 299, 5,049 and 9,535), and weight-space measures are available at that resolution only, while per-step training loss was recorded at every step.
}
  \label{tab:provenance}
  \centering
  \footnotesize
\begin{tabular}{lp{0.60\linewidth}}
\toprule
Property & Value across all 32 runs \\
\midrule
Runs & 32 \\
Distinct commits & \texttt{d52a2b8} \\
Distinct branches & \texttt{arm-cut-2026-08-08} \\
Runs with a dirty working tree & 0 \\
Runs resumed after interruption & 0 \\
Machines & \texttt{instance-00m4vv3q-main} (16), \texttt{instance-6ltvgvpi-main} (16) \\
Steps run (all runs) & [0, 9536] \\
Checkpoints written per run & 191 \\
Precision & \texttt{bf16} \\
Micro-batch $\times$ accumulation & 8$\times$32 \\
AdamW implementation & \texttt{foreach} \\
Python & \texttt{3.12.13} \\
Wall clock per run (s): min / mean / max & 35375 / 35429 / 35483 \\
\bottomrule
\end{tabular}
{}
\end{table}

\section{Robustness of the Evaluation
}
\label{app:robustness}

These are two checks on whether the dependent variable behaves as intended:
the barrier recomputed on four times as many held-out windows for a subset of
pairs (Table~\ref{tab:windows}), and a recomputation of a previously published quantity on different hardware (Table~\ref{tab:replication}). Both columns were produced by \texttt{arm\_pair\_metrics.py}, the published column on an NVIDIA A100 under \texttt{torch 2.13.0+cu130} and the recomputed column on an NVIDIA RTX A6000 under \texttt{torch 2.13.0+cu126}.
\begin{table}[h]
  \caption{
  Each row recomputes one pair's barrier on 2,048 held-out windows against the 512 held-out windows used throughout on the same alpha grid, and the final row gives the largest and mean absolute relative shift. This supports our null as a valid finding, as quadrupling the evaluation text moves every barrier by about one percent.}
  \label{tab:windows}
  \centering
  \small
\begin{tabular}{lrrr}
\toprule
Pair & \shortstack{Barrier at \\ 512 windows} & \shortstack{Barrier at \\ 2048 windows} & \shortstack{Relative \\ shift (\%)} \\
\midrule
\texttt{armtwin\_seed00\_fluent-fabricated} & 0.141220 & 0.139366 & -1.31 \\
\texttt{armtwin\_seed00\_fluent-attested} & 0.147930 & 0.145989 & -1.31 \\
\texttt{armtwin\_seed03\_fluent-fabricated} & 0.139241 & 0.139523 & +0.20 \\
\texttt{armtwin\_seed03\_fluent-attested} & 0.142782 & 0.142944 & +0.11 \\
\texttt{armtwin\_seed06\_fluent-fabricated} & 0.161798 & 0.161075 & -0.45 \\
\texttt{armtwin\_seed06\_fluent-attested} & 0.156373 & 0.155146 & -0.78 \\
\texttt{twintwin\_seed00\_seed01} & 4.811980 & 4.839452 & +0.57 \\
\texttt{twintwin\_seed02\_seed05} & 4.746679 & 4.789661 & +0.91 \\
\midrule
\textit{max $|$shift$|$ / mean $|$shift$|$} & --- & --- & 1.31 / 0.71 \\
\bottomrule
\end{tabular}
{}
\end{table}

\begin{table}[h]
  \caption{
  The published column was produced on one machine and the recomputed column on different hardware, and the final column records whether the weight distance agrees exactly. This verifies that the calculation reproduces across hardware, agreeing to about $10^{-8}$, with the largest disagreement $1.4\times10^{-8}$ against barriers around 0.12.}
  \label{tab:replication}
  \centering
  \footnotesize
\begin{tabular}{rrrrrc}
\toprule
Seed & \shortstack{Published \\ barrier} & \shortstack{Recomputed \\ barrier} & $\Delta$ & \shortstack{Published \\ Euclidean distance} & \shortstack{Euclidean distance \\ identical} \\
\midrule
0 & 0.10834173 & 0.10834172 & $-1.33\times10^{-8}$ & 220.082740 & \texttt{Yes} \\
1 & 0.10786739 & 0.10786741 & $1.39\times10^{-8}$ & 217.702401 & \texttt{Yes} \\
2 & 0.11001746 & 0.11001747 & $6.34\times10^{-9}$ & 223.639063 & \texttt{Yes} \\
3 & 0.14102024 & 0.14102025 & $7.79\times10^{-9}$ & 231.629570 & \texttt{Yes} \\
4 & 0.16180006 & 0.16180005 & $-1.08\times10^{-8}$ & 233.122265 & \texttt{Yes} \\
5 & 0.13635551 & 0.13635552 & $1.40\times10^{-8}$ & 235.610437 & \texttt{Yes} \\
6 & 0.09282110 & 0.09282110 & $2.87\times10^{-9}$ & 214.528739 & \texttt{Yes} \\
7 & 0.12402080 & 0.12402079 & $-7.35\times10^{-9}$ & 225.508019 & \texttt{Yes} \\
\midrule
\textit{mean} & 0.12278054 & 0.12278054 & $1.67\times10^{-9}$ & 225.227904 & \texttt{Yes} \\
\textit{max $|\Delta|$} & --- & --- & $1.40\times10^{-8}$ & --- & --- \\
\bottomrule
\end{tabular}
{}
\end{table}

\section{Pre-registration
}
\label{app:prereg}

All 32 runs were trained from d52a2b8, and every row of Table~\ref{tab:prereg} is a date tested against that commit. The repository is public at \url{https://github.com/zacharyspeck/burst-study}, and the file and line references in that table are as of the commit cited rather than as of the current head. The conditions are named \texttt{fluent-attested} and \texttt{fluent-fabricated} in current code and outputs, and \texttt{fluent-true} and \texttt{fluent-false} in the dated records, including the pre-registration, which retain the names in use when each decision was fixed. No confirmatory-arm run existed at any seed when the registered items were fixed, so no registered contrast's outcome was available to choose by. The contrast was registered under the truth framing and the attestation framing is a reinterpretation adopted after the result was known, with the registered contrast, metric, test, and correction policy unchanged.
\begin{table}[h]
  \caption{The commit column is the commit in which each decision was recorded, the
fourth column tests each date against the commit all 32 runs were trained
from, and the final column gives the file and line where the
decision is written down.
  }
  \label{tab:prereg}
  \centering
  \small
  \setlength{\tabcolsep}{4pt}
\begin{tabular}{>{\raggedright\arraybackslash}p{0.27\linewidth}llc>{\raggedright\arraybackslash}p{0.30\linewidth}}
\toprule
What was fixed & Date & Commit & \shortstack{Before the \\ 32 runs?} & Recorded at \\
\midrule
Primary outcome measure & 2026-08-03 & \texttt{2ef812b} & \texttt{Yes} & \texttt{docs/\allowbreak{}preregistration.\allowbreak{}md:\allowbreak{}174-\allowbreak{}-\allowbreak{}222} \\
Metric decision rule (plain vs.\ aligned barrier) & 2026-08-03 & \texttt{2ef812b} & \texttt{Yes} & \texttt{docs/\allowbreak{}preregistration.\allowbreak{}md:\allowbreak{}202-\allowbreak{}-\allowbreak{}222} \\
Branch that rule selected: plain barrier & 2026-08-06 & \texttt{715dd67} & \texttt{Yes} & \texttt{docs/\allowbreak{}measurements/\allowbreak{}12-\allowbreak{}injection-\allowbreak{}point-\allowbreak{}ruler.\allowbreak{}md} \\
Primary contrast & 2026-08-03 & \texttt{42a7d35} & \texttt{Yes} & \texttt{docs/\allowbreak{}preregistration.\allowbreak{}md:\allowbreak{}90-\allowbreak{}-\allowbreak{}122} \\
Secondary contrast, registered & 2026-08-03 & \texttt{42a7d35} & \texttt{Yes} & \texttt{docs/\allowbreak{}preregistration.\allowbreak{}md:\allowbreak{}124-\allowbreak{}-\allowbreak{}148} \\
Secondary contrast, cut with its condition & 2026-08-08 & \texttt{8d3ae2a} & \texttt{Yes} & \texttt{docs/\allowbreak{}preregistration.\allowbreak{}md:\allowbreak{}392-\allowbreak{}-\allowbreak{}433} \\
Correction policy: none, family 1 & 2026-08-08 & \texttt{1ea243b} & \texttt{Yes} & \texttt{docs/\allowbreak{}decisions-\allowbreak{}pending.\allowbreak{}md:\allowbreak{}704-\allowbreak{}-\allowbreak{}718} \\
Condition set: seven conditions & 2026-08-03 & \texttt{c2df6c7} & \texttt{Yes} & \texttt{burst/\allowbreak{}config.\allowbreak{}py:\allowbreak{}72-\allowbreak{}-\allowbreak{}80} \\
Condition set: cut to four & 2026-08-08 & \texttt{8d3ae2a} & \texttt{Yes} & \texttt{burst/\allowbreak{}config.\allowbreak{}py:\allowbreak{}88-\allowbreak{}-\allowbreak{}95} \\
\bottomrule
\end{tabular}
{}
\end{table}

\end{document}